\documentclass{article}
\usepackage{spconf,amsmath,graphicx}
\usepackage{booktabs}
\usepackage{amsfonts,amssymb}
\usepackage{bbm}
\usepackage{array,multirow}
\usepackage{threeparttable}
\usepackage{adjustbox}
\usepackage{colortbl}
\usepackage{graphicx}
\usepackage{booktabs}
\usepackage{subfigure} 
\usepackage{url}
\usepackage[dvipsnames]{xcolor}

\title{Dynamic Dual Sampling Module for Fine-Grained Semantic Segmentation}
\name{Chen Shi$^{1}$$^+$, Xiangtai Li$^{2}$$^+$, Yanran Wu$^{1}$, Yunhai Tong$^{2}$, Yi Xu$^{1}$\sthanks{Corresponding author. $^+$ The first two authors contribute equally. This study was partially supported by 111 project (BP0719010), Shanghai Science and Technology Committee (18DZ2270700).}}
\address{$^{1}$Shanghai Jiao Tong University, China \\
        $^{2}$ Peking University, China }
\begin{document}
%
\maketitle

\begin{abstract}
Representation of semantic context and local details is the essential issue for building modern semantic segmentation models. However, the interrelationship between semantic context and local details is not well explored in previous works. In this paper, we propose a Dynamic Dual Sampling Module (DDSM) to conduct dynamic affinity modeling and propagate semantic context to local details, which yields a more discriminative representation. Specifically, a dynamic sampling strategy is used to sparsely sample representative pixels and channels in the higher layer, forming adaptive compact support for each pixel and channel in the lower layer. The sampled features with high semantics are aggregated according to the affinities and then propagated to detailed lower-layer features, leading to a fine-grained segmentation result with well-preserved boundaries. Experiment results on both Cityscapes and Camvid datasets validate the effectiveness and efficiency of the proposed approach.
Code and models will be available at \url{x3https://github.com/Fantasticarl/DDSM}.

\end{abstract}

\section{Introduction}
\label{sec:intro}
Semantic segmentation, which entails assigning a label to each pixel of an image, is useful in a growing number of applications,
including augmented reality, surveillance, and autonomous
driving. With the development of deep FCN networks~\cite{fcn,pspnet,DAnet}, the related works mainly focus on two aspects: global context modeling~\cite{pspnet,deeplabv3} and local details modeling~\cite{xiangtl_gff}. The former models the long-range dependencies among pixels on the higher level of the network by overcoming the limited receptive field of the convolution network. The latter imports extra components such as lower-level features~\cite{deeplabv3p} or includes edge supervision~\cite{xiangtl_decouple} for finer and detailed results. The feature pyramids encode different scaled features where the higher layers contain coarse semantics while the lower layers represent fine details~\cite{fpn,sfnet}. However, the interrelationship between semantic context and local details is not well explored. In this paper, we focus on exploring the interrelationship between two different layers. Since the semantic gaps~\cite{ding_context}, our solution enhances lower-layer features based on its affinity with the higher layer instead of directly adding features in FPN~\cite{fpn}, successfully propagating the semantic context to local details via dynamic sampling of representative pixels and channels in higher layers. 

\begin{figure}[!t]
\begin{minipage}[b]{0.32\linewidth}
  \centering
  \centerline{\includegraphics[width=\linewidth]{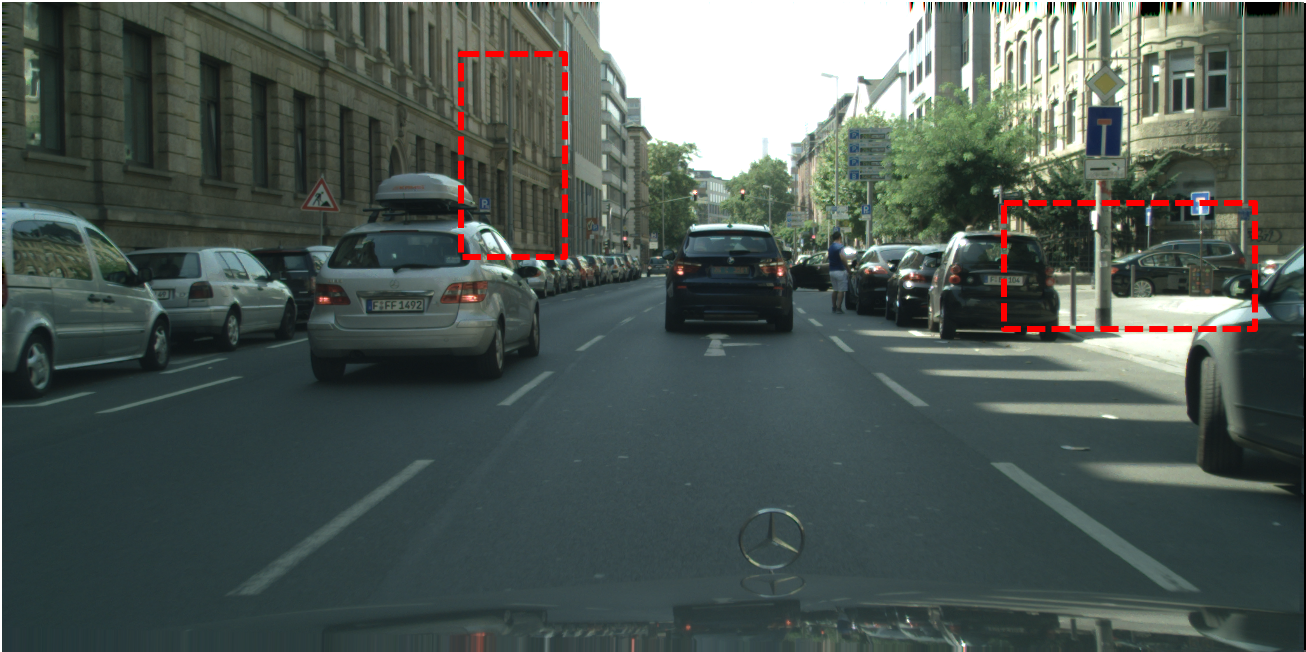}}
  \centerline{(a)}\medskip
\end{minipage}
\hfill
\begin{minipage}[b]{0.32\linewidth}
  \centering
  \centerline{\includegraphics[width=\linewidth]{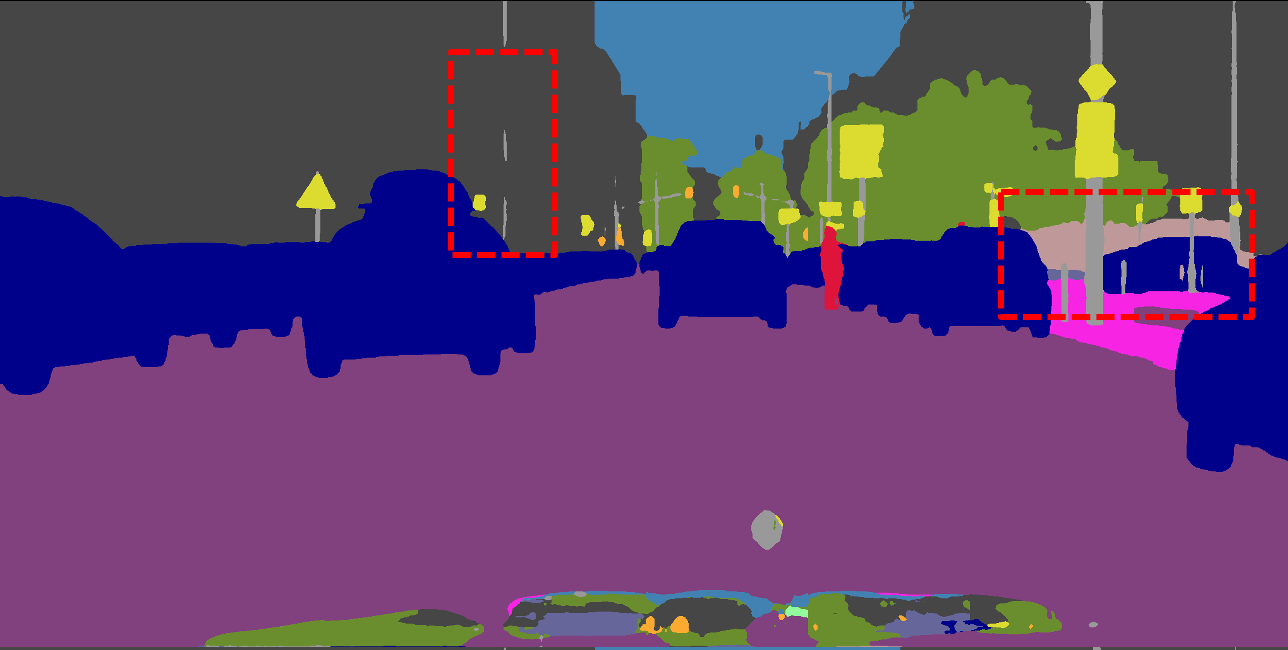}}
  \centerline{(c)}\medskip
\end{minipage}
\hfill
\begin{minipage}[b]{0.32\linewidth}
  \centering
  \centerline{\includegraphics[width=\linewidth]{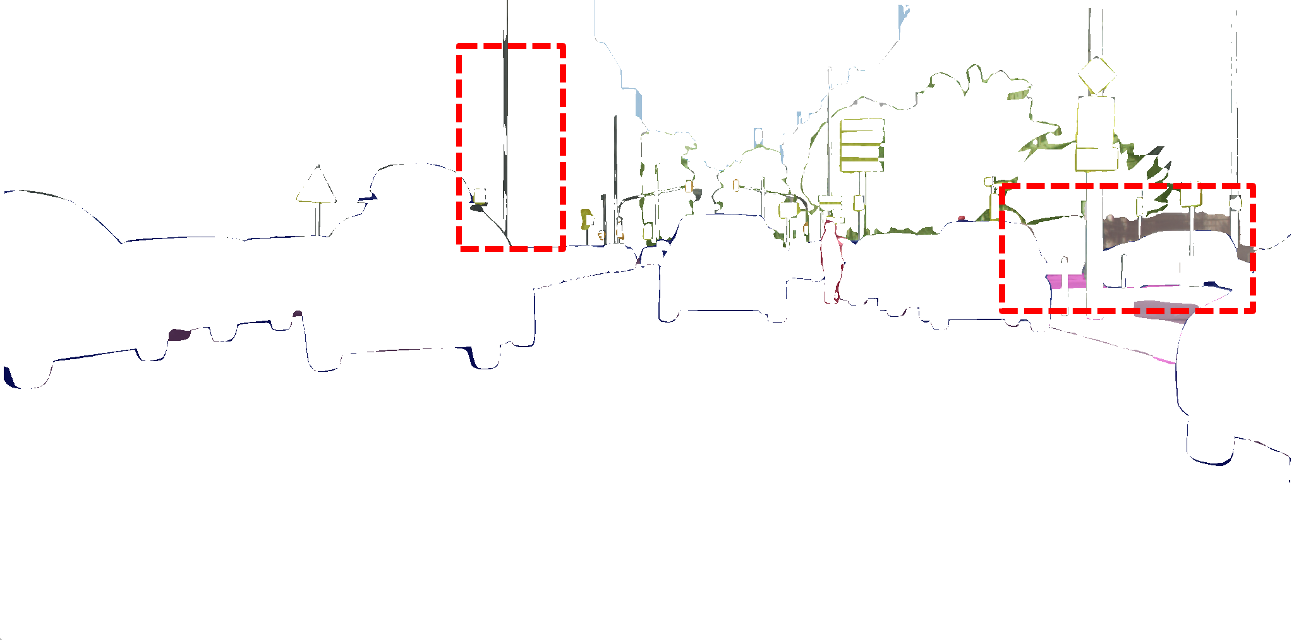}}
  \centerline{(e)}\medskip
\end{minipage} \\
\begin{minipage}[b]{0.32\linewidth}
  \centering
  \centerline{\includegraphics[width=\linewidth]{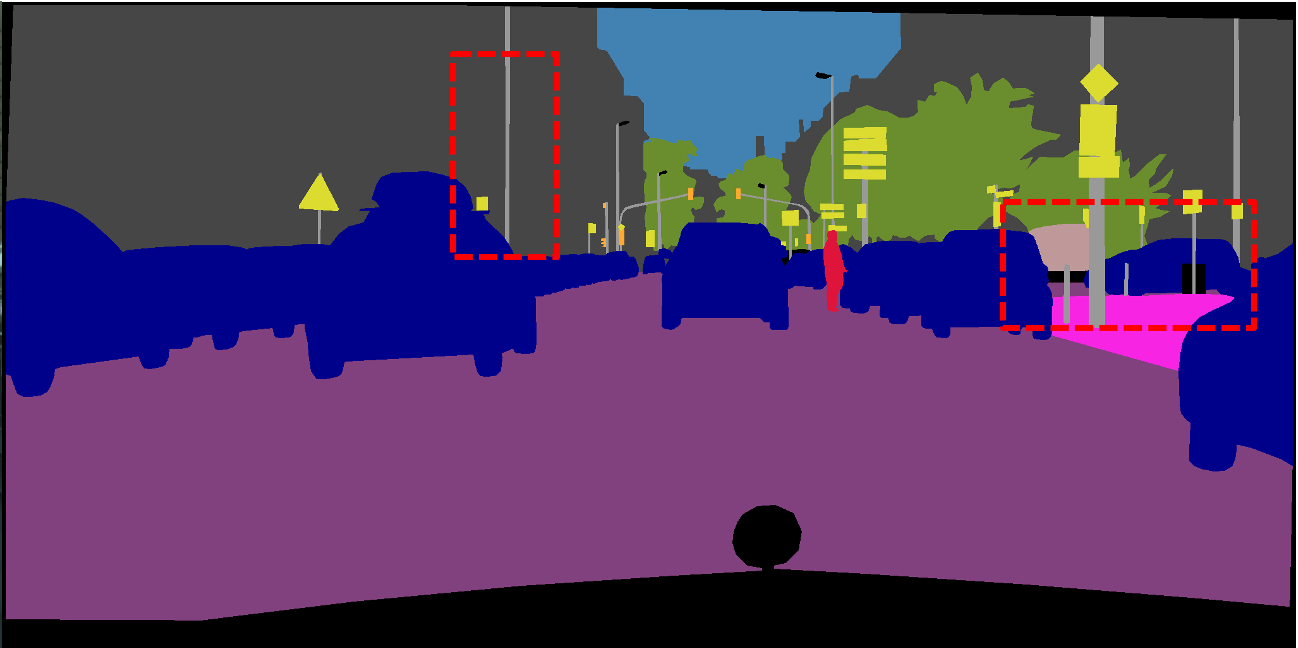}}
  \centerline{(b)}\medskip
\end{minipage}
\hfill
\begin{minipage}[b]{0.32\linewidth}
  \centering
  \centerline{\includegraphics[width=\linewidth]{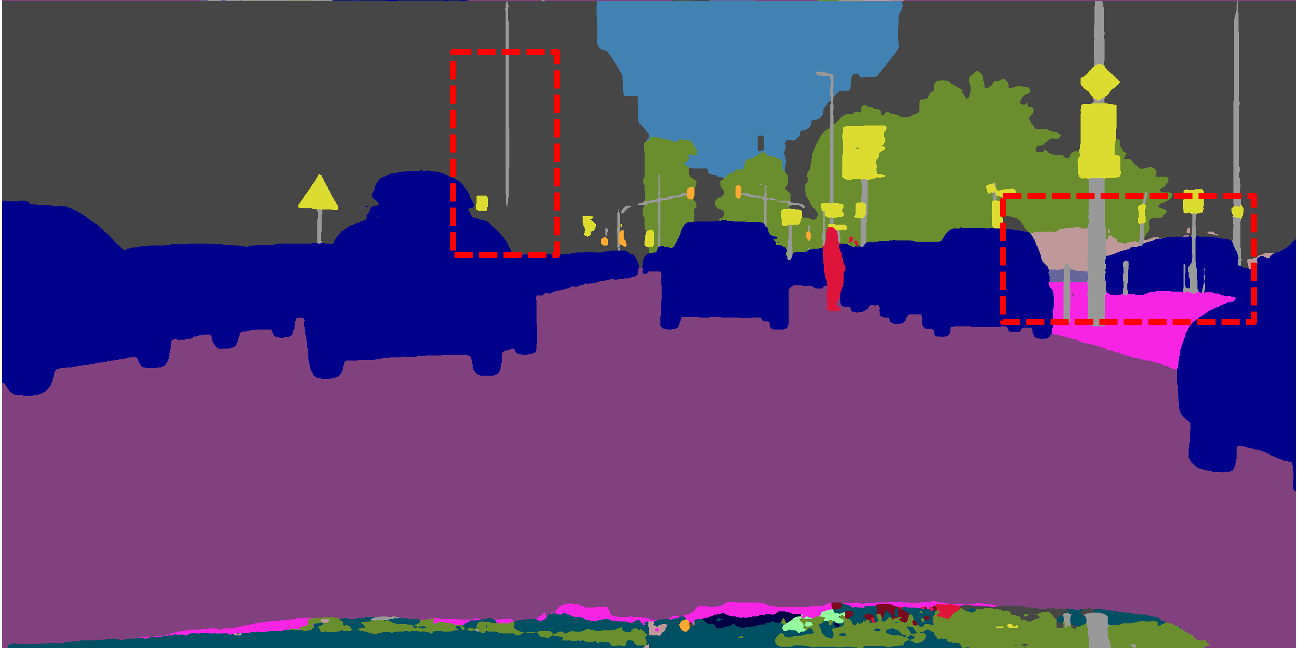}}
  \centerline{(d)}\medskip
\end{minipage}
\hfill
\begin{minipage}[b]{0.32\linewidth}
  \centering
  \centerline{\includegraphics[width=\linewidth]{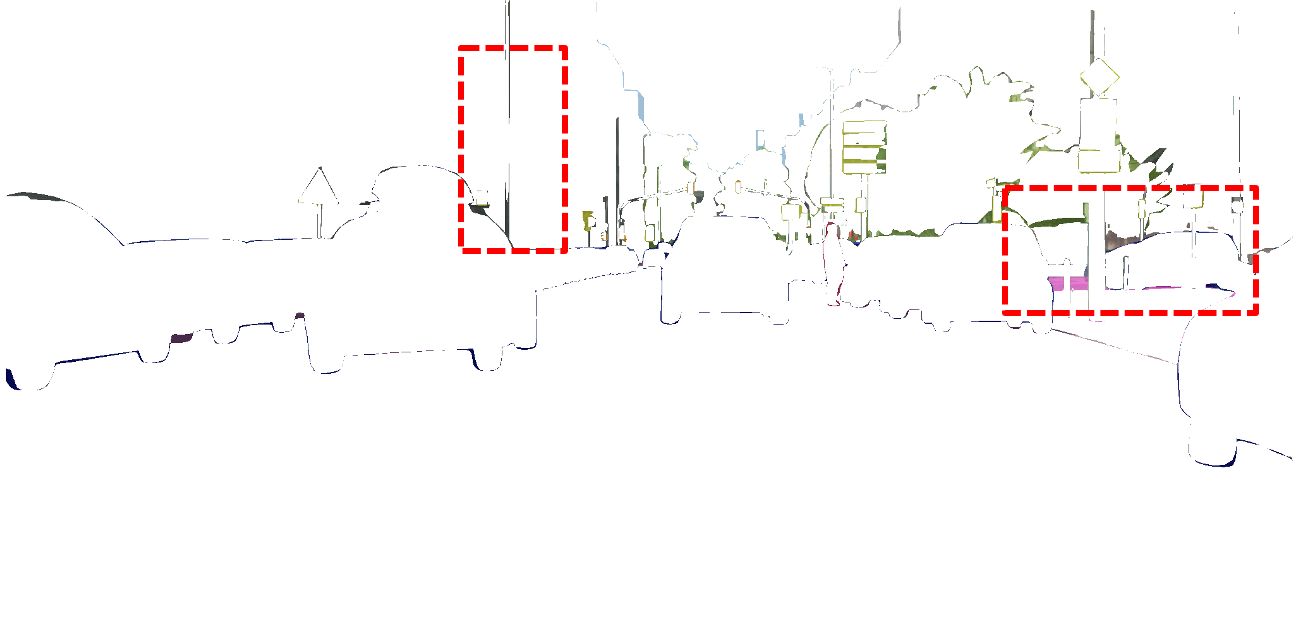}}
  \centerline{(f)}\medskip
\end{minipage}
\caption{\textbf{Visualizations of the prediction and error map.} Our model outputs finer boundaries and small objects. (a) Input, (b) Ground truth, (c) Output of UPerNet~\cite{upernet}, (d) Output of ours, (e) Error map of UPerNet~\cite{upernet}, (f) Error map of Ours.} 
\label{fig:vis}
\end{figure}

It is noted that the existing affinity modeling methods, including self-attention based~\cite{a2net} or graph-based~\cite{beyond_grids} models, require expensive pixel-wised computation across the whole image. For instance, FPT~\cite{zhang2020feature} uses a transformer to model the adjacent features' affinity, leading to immense resource cost. There is a rather high redundancy since the natural image meets the piece-wise smoothness constraint that the pixels within the same segment share certain visual characteristics. Accordingly, inspired by dynamic graph modeling~\cite{Zhang_2020_CVPR,Yu-ECCV-RepGraph-2020} and deformable convolutions~\cite{deformable,zhu2019deformable}, we propose a dynamic affinity modeling method to avoid redundancy and achieve efficient feature propagation. Rather than using full pixels, we sample representative pixels to form adaptive compact support. Furthermore, we propose a dynamic sampler based on DCNv2~\cite{zhu2019deformable} instead of sampling fixed neighborhood pixels to fit the orientation distribution of image structures. Moreover, the channel encodes corresponding class-specific information, and several works~\cite{DAnet} have shown the advantages of considering spatial and channel simultaneously to enhance class-aware information in feature representation. Hence, we also apply the dynamic sampler in feature channels and conduct channel-wise dynamic affinity modeling. After obtaining the sampled pixels/channels, we calculate both spatial and channel-wise affinities for each pixel/channel in the lower-layer. Finally, relevant semantics is aggregated according to the affinities and propagated to detailed lower-layer features.

In summary, we propose a Dynamic Dual Sampling Module(DDSM), which contains spatial-wise and channel-wise dynamic affinity modeling. The representation of the pixels /channels in the lower layer is enhanced by some dynamically sampled pixels/channels from the higher layer. We validate DDSM's effectiveness on two typical networks, UPerNet~\cite{upernet} and Deeplabv3+~\cite{deeplabv3p}. Notably, our method could provide fine-grained segmentation results with well-preserved boundaries. Fig.~\ref{fig:vis} shows the error map refined by DDSM based on UPerNet~\cite{upernet}. After inserting DDSM into UPerNet~\cite{upernet}, it achieves competitive results on Cityscapes~\cite{Cityscapes} and Camvid~\cite{CamVid}. In particular, DDSM outperforms DAnet~\cite{DAnet} with only 30\% computation during the inference.


\section{Method}
\label{Method}

In this section, we first describe our dynamic sampler, which is inspired by DCNv2~\cite{zhu2019deformable}. Then, we give a detailed introduction of our proposed Dynamic dual sampling module, which dynamically samples pixels and channels simultaneously. Finally, we deploy our proposed module into two frameworks.

\subsection{Dynamic Sampler}
\label{sec:DS}
\begin{figure}[!t]
    \centering
    \includegraphics[width=0.6\linewidth]{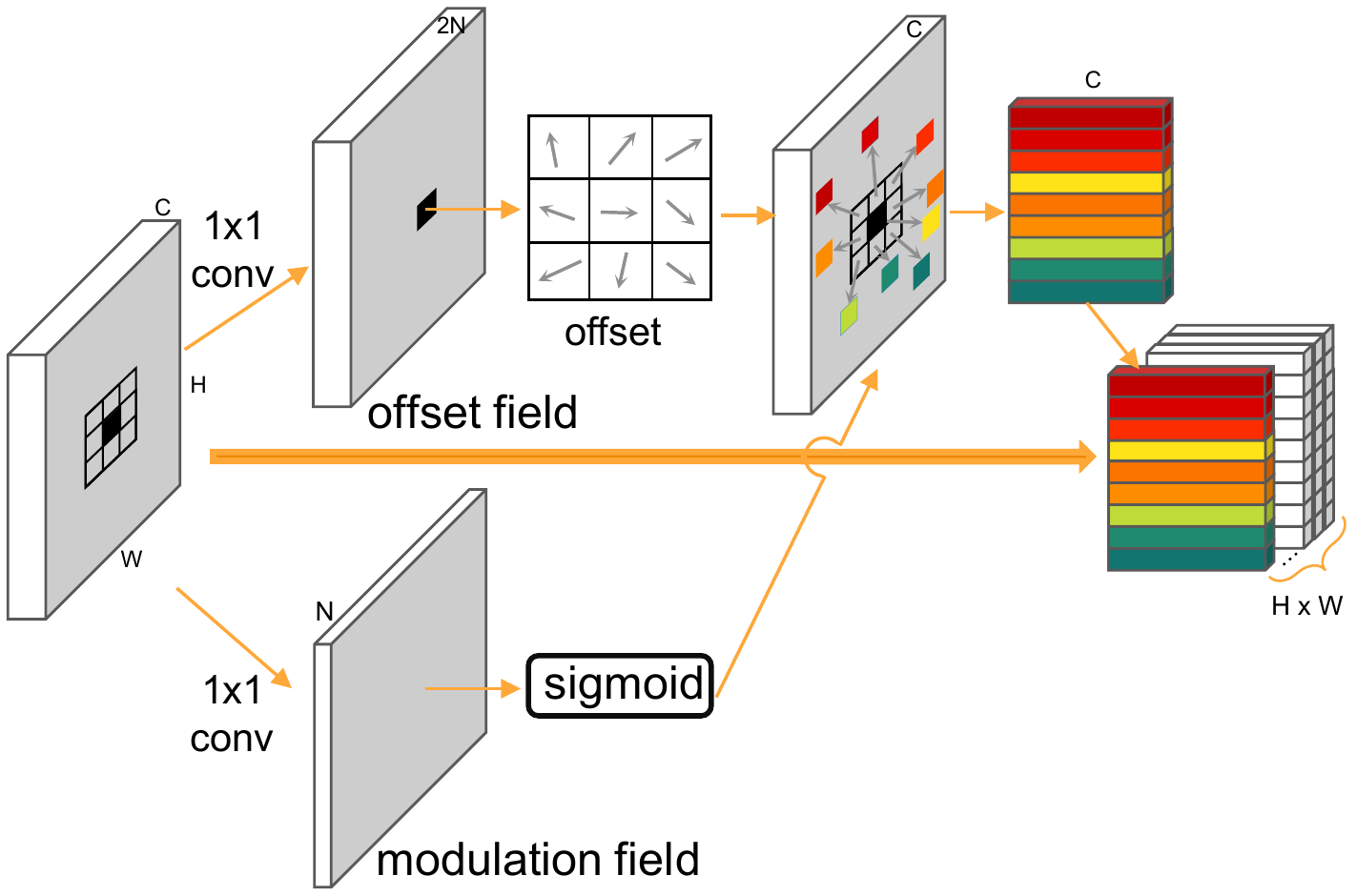}
    \caption{\textbf{Dynamic sampler based on DCN-v2}~\cite{zhu2019deformable}. It samples representative positions with learned offsets and modulations. }
    \label{fig:DS}
\end{figure}

Given the input features $\mathbf{x}$ with dimension $C \times H \times W$, the sampler dynamically samples $N = k \times k$ pixels from $\mathbf{x}$ for each position $p$, as Fig.~\ref{fig:DS} shows. Specifically, a regular grid $\mathcal{R}_{k \times k}=\{p_n|n=1,2,...,N\}$ is defined to get an initial sampling area of $p$. Then, we use a $1 \times 1$ convolution layer instead of $3 \times 3$ in DCN~\cite{deformable} to learn an offset for each position in the grid $\mathcal{R}_{k \times k}$, and then an offset map with dimension of $2N \times H \times W$ is obtained, in which $N$ 2D offsets $\Delta p_{n} = (q_x, q_y), n=1,2, \cdots, N$ are learned. With the offset map, we use bilinear interpolation to compute the sampled features $\mathbf{x}(p+p_{n}+\Delta p_{n})$ of each sampled position $p+p_{n}+\Delta p_{n}$. To learn the offset map more flexibly and further boost the performance, a learnable scalar $\Delta m_n$ is added following the work of DCNv2~\cite{zhu2019deformable}. Given the dynamic sampler $\mathbf{F}$, the sampled features can be formulated as Eq.\ref{equ:sampler}:
\begin{equation}
\label{equ:sampler}
\mathbf{F}(\mathbf{x}(p))=\{\mathbf{x}(p+p_n+\Delta p_{n})\Delta m_n|n=1,2,..., N \}.
\end{equation}
The output ($C \times H \times W \times N$) gives the features at $N$ positions sampled from $\mathbf{x}$ for each position $p$ in the $H \times W$ feature map.

\subsection{Dynamic Dual Sampling Module(DDSM)}

\begin{figure*}[htb]
\begin{minipage}[b]{.48\linewidth}
  \centering
  \centerline{\includegraphics[width=\linewidth]{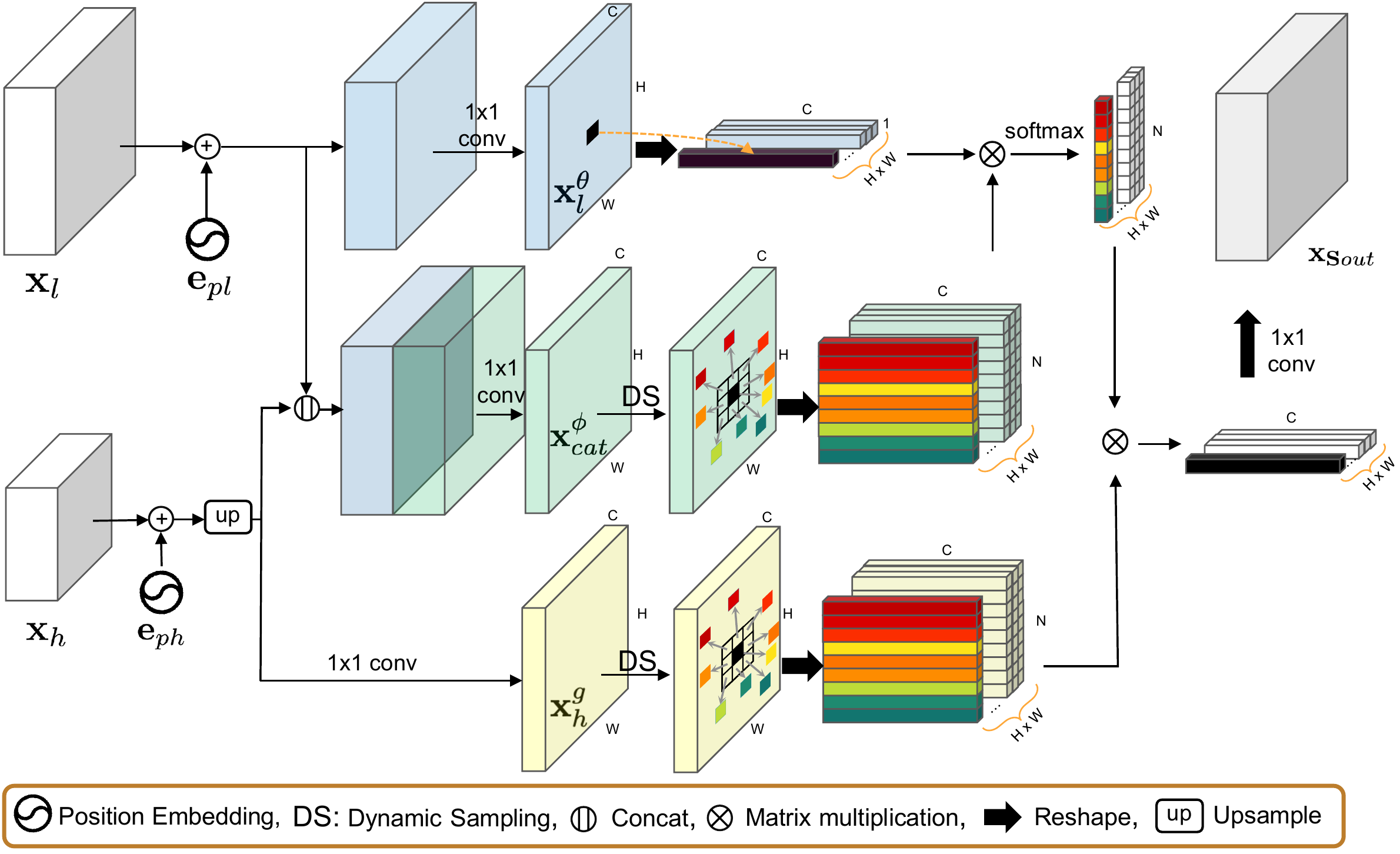}}

  \centerline{(a) Spatial-wise dynamic affinity modeling}\medskip
\end{minipage}
\hfill
\begin{minipage}[b]{0.48\linewidth}
  \centering
  \centerline{\includegraphics[width=\linewidth]{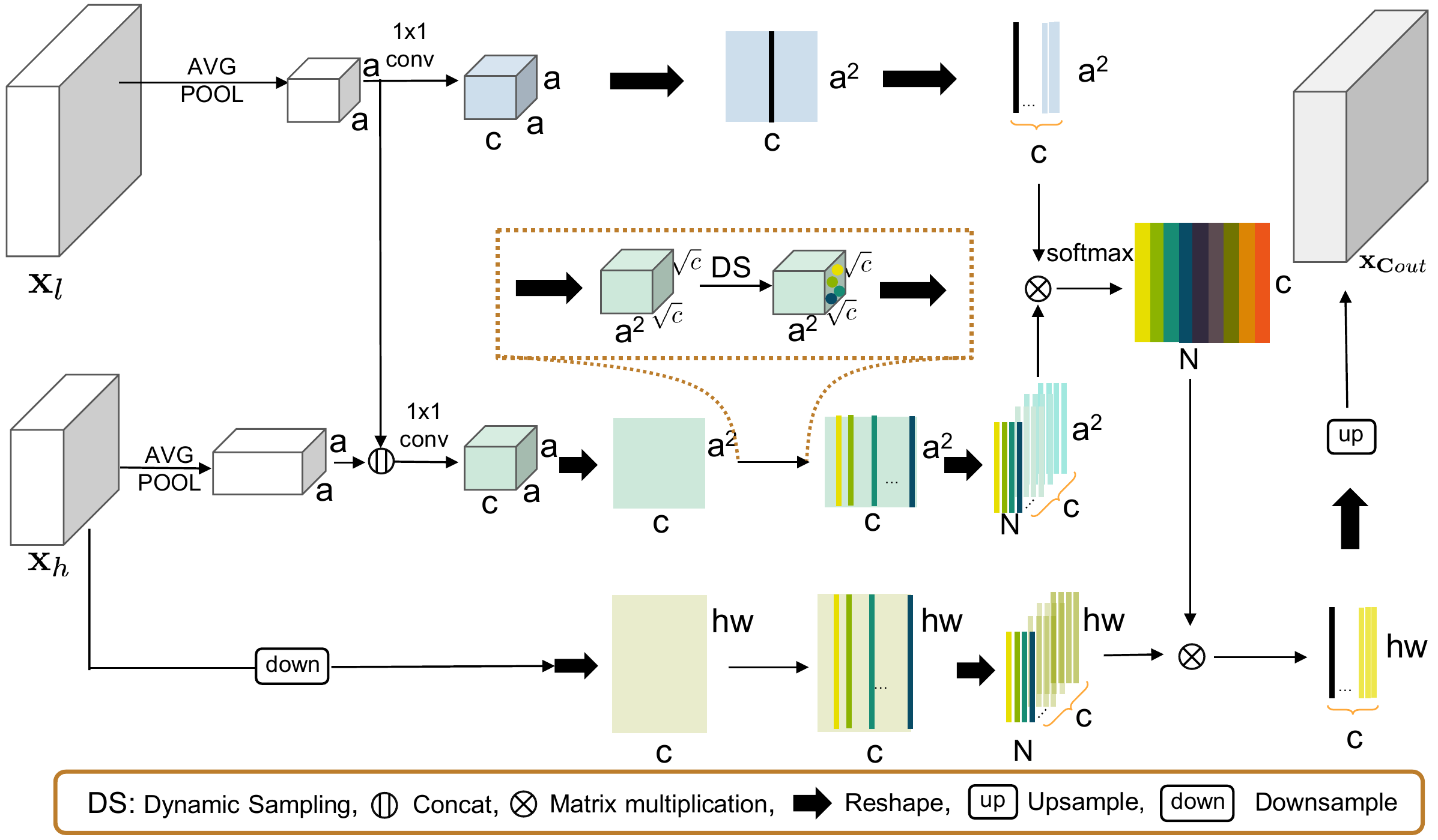}}
  \centerline{(b) Channel-wise dynamic affinity modeling}\medskip
\end{minipage}

\caption{\textbf{Dynamic Dual Sampling Module(DDSM).} DDSM contains two parts, spatial-wise dynamic affinity modeling and channel-wise dynamic affinity modeling. The two parts accept the same inputs of two different features. Note that $\mathbf{x}_l$ and $\mathbf{x}_h$ need to be reduced to the same dimension with $1 \times 1$ Conv in advance. The final output of the module will be $\mathbf{x}_l+\mathbf{x}_{\mathbf{S}out}+\mathbf{x}_{\mathbf{C}out}$.}
\label{fig:DDSM}

\end{figure*}

The Dynamic Dual Sampling Module consists of two parts, namely spatial-wise dynamic affinity modeling and channel-wise dynamic affinity modeling. The final output is the summation of the output features from both parts.

\noindent
\textbf{Spatial-wise Dynamic Affinity Modeling:} This part dynamically assigns features of $N$ pixels sampled from the higher layer for each pixel in the lower layer. As shown in Fig.~\ref{fig:DDSM}(a), for the low-level features $\mathbf{x}_l$ and the high-level features $\mathbf{x}_h$, we first upsample $\mathbf{x}_h$ to $H \times W$, same as $\mathbf{x}_l$. Note that position information is crucial in feature fusion. A common method of introducing position information is to summarize features and positional encodings as input~\cite{DETR}. We add learnable positional embeddings~\cite{DETR} $\mathbf{e}_{pl}$, $\mathbf{e}_{ph}$ to the features $\mathbf{x}_l$, $\mathbf{\hat x}_h$ to disambiguate different spatial positions. 
Then we concatenate both $\mathbf{x}_h$ and $\mathbf{x}_l$ into features $ \mathbf{x}_{cat}=(\mathbf{x}_l+\mathbf{e}_{pl})||(\mathbf{\hat x}_h+\mathbf{e}_{ph})$. Then, we use three $1 \times 1$ convolution layers to do dimension reduction on $\mathbf{x}_{cat}$, $\mathbf{x}_l+\mathbf{e}_{pl}$, and $\mathbf{\hat x}_h+\mathbf{e}_{ph}$, forming a new feature set as
$\mathbf{x}_l^{\theta} = W_{\theta}( \mathbf{x}_l+\mathbf{e}_{pl})$, 
$\mathbf{x}_{cat}^{\phi} = W_{\phi}\mathbf{x}_{cat}$,
$\mathbf{x}_h^{g} = W_{g} (\mathbf{\hat x}_h+\mathbf{e}_{ph})$. Similar to the definition in the work~\cite{vaswani2017attention}, $\mathbf{x}_l^{\theta}$, $\mathbf{x}_{cat}^{\phi}$ and $\mathbf{x}_h^{g}$ correspond to Query, Key, and Value function. 

The work~\cite{vaswani2017attention} uses the entire feature map to calculate the affinity map. Nevertheless, we use the dynamic sampler $\mathbf{F}$ to sample $N$ pixels in the Key for each position in the Query to obtain the affinity map. We sample $N$ pixels from $ \mathbf{x}_{cat}^{\phi}$ to form sampled features for each position $p$ in $\mathbf{x}_l^{\theta}$. Matrix multiplication $\textcolor{red}{X}^{1 \times C} \times \textcolor{blue}{X}^{C \times N}$ is performed between the features of each position in $\mathbf{x}_l^{\theta}$ and the transposed sampled features to form the affinity map with softmax normalization. Then, matrix multiplication $\textcolor{red}{X}^{1 \times N} \times \textcolor{blue}{X}^{N \times C}$ is performed between the affinity map and sampled features from $\mathbf{x}_h^{g}$. The above two processes are executed $N$ times to obtain the aggregation result of $N$ sampled features, which will be assigned to the low-level features through a summation operation. The spatial-wise dynamic affinity modeling is formulated as Eq.\ref{equ:att},
\begin{equation}\label{equ:att}
\mathbf{x}_{\mathbf{S}out}(p)=\sum\nolimits_{n=1}^{N} \delta [\mathbf{x}_l^{\theta}(p) \mathbf{F}(\mathbf{x}_{cat}^{\phi}(n))^{\top} ] \mathbf{F}(\mathbf{x}_h^{g}(n)),
\end{equation}
where $\mathbf{x}_{\mathbf{S}out}(p)$ is the augmented feature, $p$ is a position in $\mathbf{x}_l$ and $\delta$ is Softmax. Fig.~\ref{fig:DDSM}(a) gives the detailed pipline.

\noindent
\textbf{Channel-wise Dynamic Affinity Modeling:} The channel-wise dynamic affinity modeling is built to explore interdependencies along channels since the channel encodes class-specific information. Different from previous works~\cite{DAnet,senet}, our module dynamically samples the channels of high-level features and aggregates them according to affinity to enhance low-level features. The input $\mathbf{x}_l$ and $\mathbf{x}_h$ are both average pooled to $a \times a$ to reduce the spatial resolution. $\mathbf{x}_l$ and $\mathbf{x}_h$ is concatenated and then reshaped to $c \times a^2$ to perform channel-wise dynamic sampling. As Fig.~\ref{fig:DDSM}(b) shows, in order to reuse the dynamic sampler $\mathbf{F}$, we reshape the features with $c \times a^2$ dimension into $a^2 \times \sqrt{c} \times \sqrt{c}$ (we set channel $c$ as a square number 64 in this work), and then reshape back to $c \times a^2 \times N_c$ for affinity calculation after the sampling of $N_c$ channels. For each channel in the pooled low-level feature, the affinity map is calculated with the corresponding sampled $N_c$ channels. Finally, the high-level features are also dynamically sampled in the channel dimension and propagated to the corresponding channel of the low-level features according to the affinity map. Note that we downsample $\mathbf{x}_h$ to $16 \times 16$ to reduce computation here. The whole process can be formulated as Eq.~\ref{equ:catt}:
\begin{equation}\label{equ:catt}
\mathbf{x}_{\mathbf{C}out}(c)=\sum\nolimits_{n=1}^{N_c} \delta [\mathbf{x}_l^{\alpha}(c) \mathbf{F}(\mathbf{x}_{cat}^{\beta}(n))^{\top} ] \mathbf{F}(\mathbf{x}_h^{\gamma}(n)),
\end{equation}
where $c$ is the channel in low-level features,
$\mathbf{x}_l^{\alpha} = W_{\alpha} \mathbf{x}_l^p$,  
$\mathbf{x}_{cat}^{\beta} = W_{\beta}(\mathbf{x}_l^p||\mathbf{x}_h^p$),
$\mathbf{x}_h^{\gamma} = W_{\gamma} \mathbf{x}_h^d$, $\mathbf{x}_l^p$ and $\mathbf{x}_h^p$ are the pooled low-level and high-level features, $\mathbf{x}_h^d$ is the downsampled high-level features, $\delta$ means Softmax, $W_{\alpha}$,$W_{\beta}$,$W_{\gamma}$ are implemented with $1\times1$ convolution layers.

\noindent
\textbf{Plugin into Two Architectures:}
Our proposed DDSM is end-to-end trainable, and it can dynamically propagate rich semantic information between adjacent features. We use DDSM in UPerNet~\cite{upernet}, which contains FPN\cite{fpn_slide} with Pyramid Pooling Module(PPM)\cite{pspnet}, and Deeplabv3+\cite{deeplabv3p}. In UPerNet~\cite{upernet}, we replace the upsample module with the proposed DDSM. Let $\{\mathbf{x}_{s} | s=2,3,4,5\}$ be the output of each stage $s$ in encoder, e.g. ResNet~\cite{resnet}. The enhanced higher-level features $\mathbf{\widetilde x}_s$ and the corresponding $\mathbf{x}_{s-1}$ are passed into DDSM to form their enhanced bottom level features $\mathbf{\widetilde x}_{s-1}$. For Deeplabv3+~\cite{deeplabv3p}, we insert one DDSM module which dynamically assigns the output of ASPP $aspp(\mathbf{x}_{5})$ to the low-level $\mathbf{x}_{2}$ to form $\mathbf{\widetilde x}_{2}$ for final segmentation. 
 
\section{Experiment}
\label{exp}

\begin{figure*}[!t]
	\scalebox{0.80}{
\begin{minipage}[b]{.2\linewidth}
  \centering
  \centerline{\includegraphics[width=\linewidth]{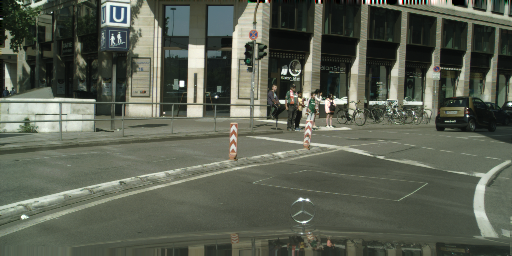}}
\end{minipage}
\begin{minipage}[b]{.2\linewidth}
  \centering
  \centerline{\includegraphics[width=\linewidth]{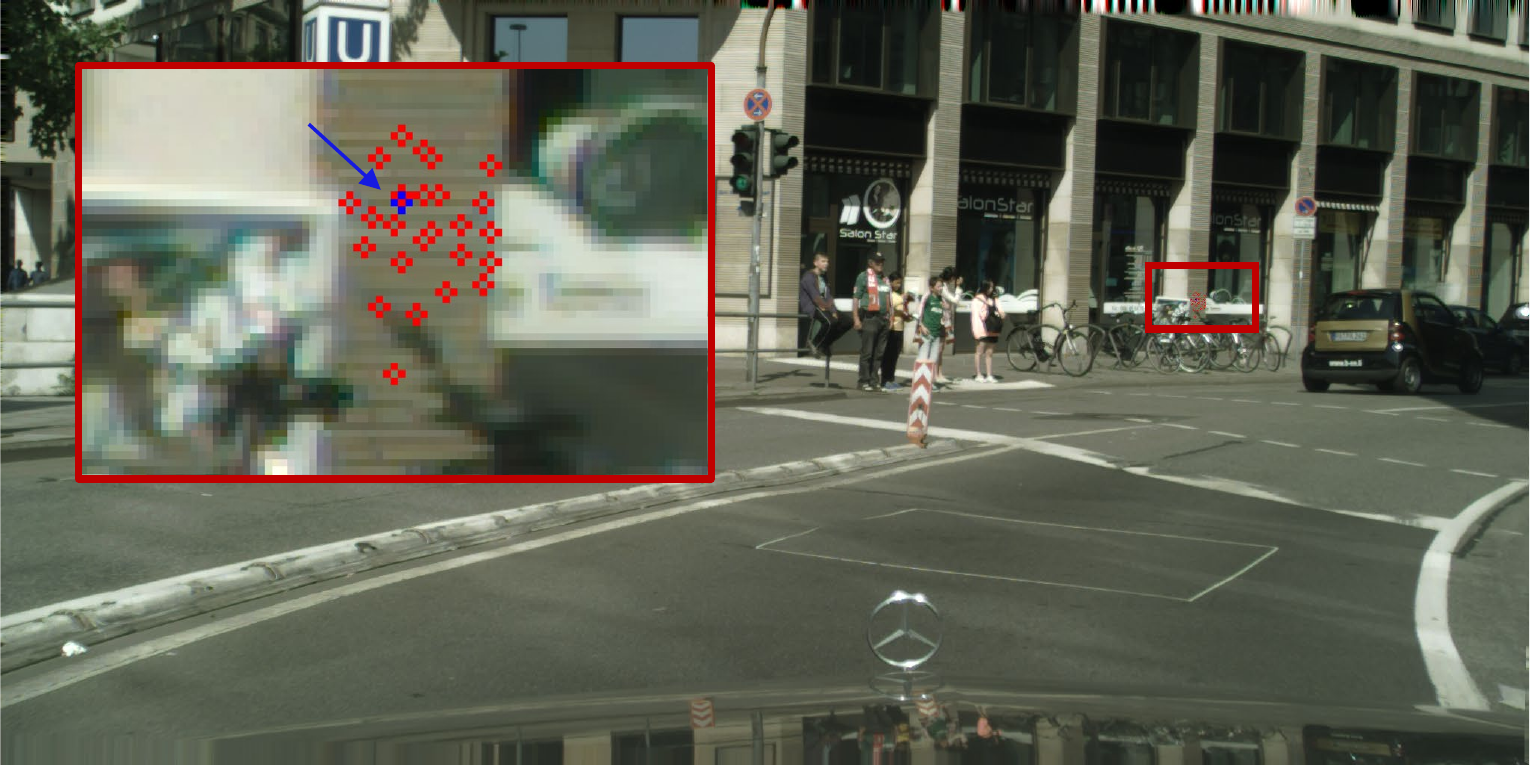}}
\end{minipage}
\begin{minipage}[b]{.2\linewidth}
  \centering
  \centerline{\includegraphics[width=\linewidth]{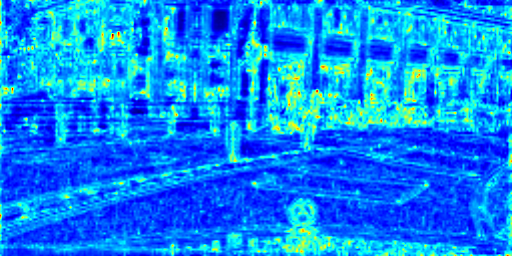}}
\end{minipage}
\begin{minipage}[b]{.2\linewidth}
  \centering
  \centerline{\includegraphics[width=\linewidth]{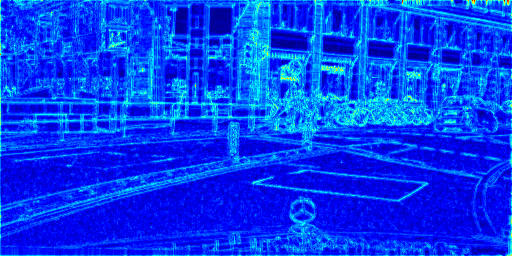}}

\end{minipage}
\begin{minipage}[b]{.2\linewidth}
  \centering
  \centerline{\includegraphics[width=\linewidth]{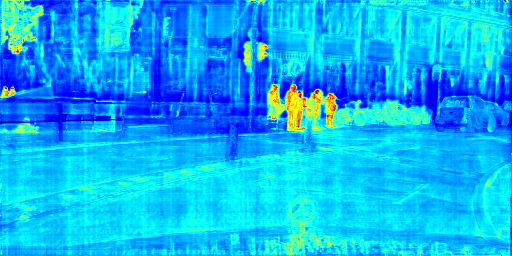}}
\end{minipage}
\begin{minipage}[b]{.2\linewidth}
  \centering
  \centerline{\includegraphics[width=\linewidth]{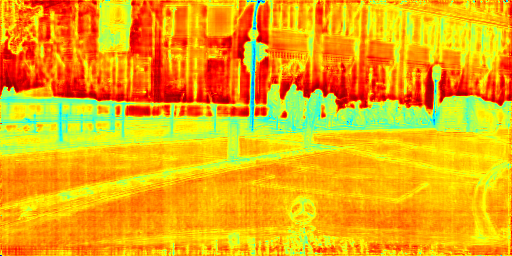}}
\end{minipage}
}
\\
\scalebox{0.80}{
\begin{minipage}[b]{0.2\linewidth}
  \centering
  \centerline{\includegraphics[width=\linewidth]{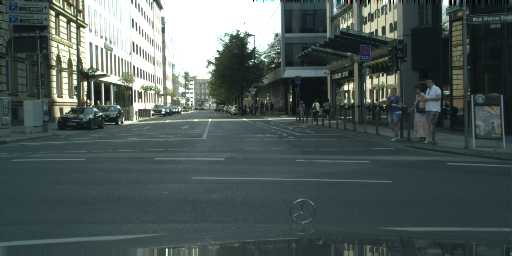}}
  \centerline{(a) Input}\medskip
\end{minipage}
\begin{minipage}[b]{0.2\linewidth}
  \centering
  \centerline{\includegraphics[width=\linewidth]{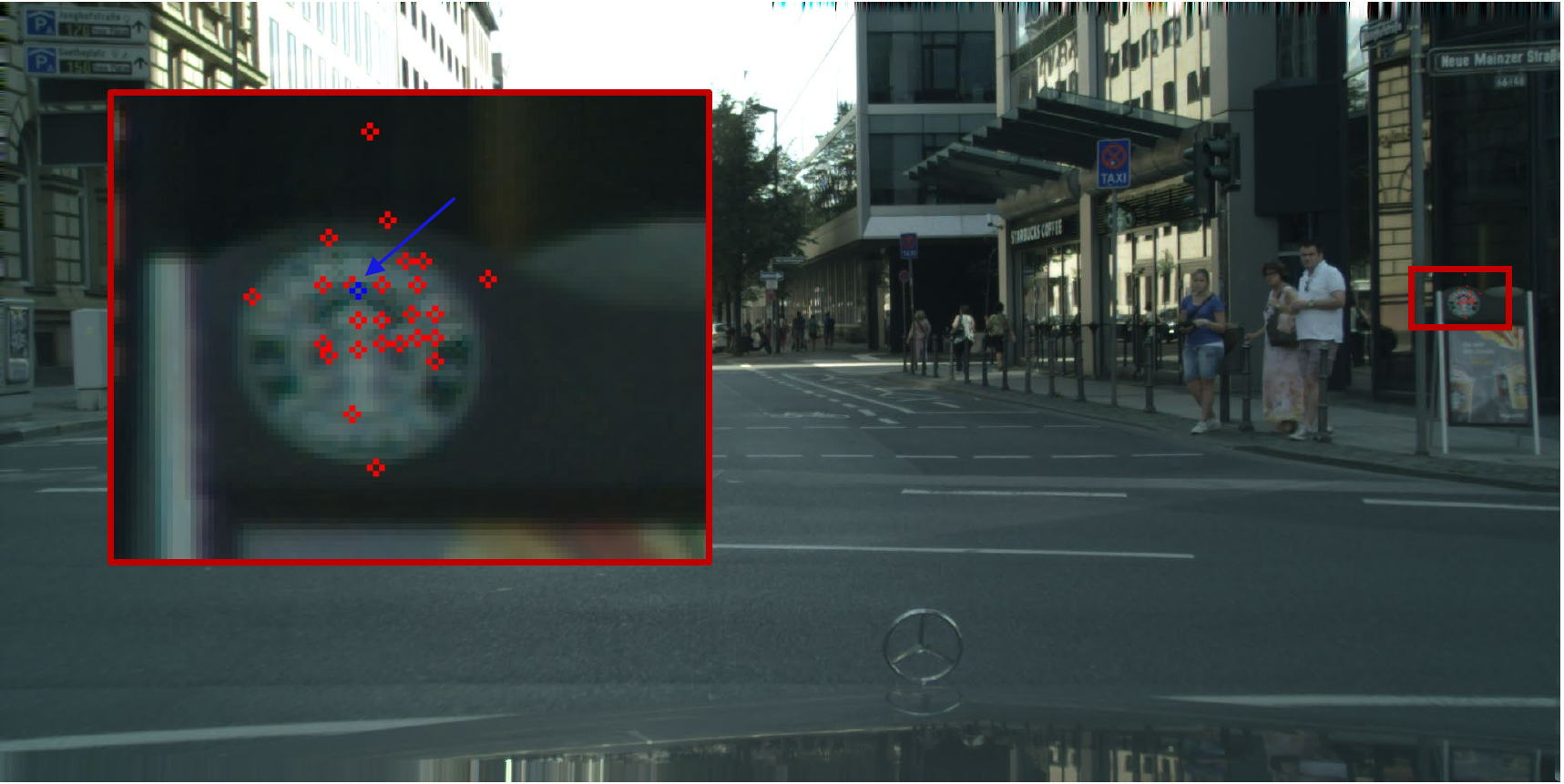}}
  \centerline{(b) $p_{sampled}$}\medskip
\end{minipage}
\begin{minipage}[b]{0.2\linewidth}
  \centering
  \centerline{\includegraphics[width=\linewidth]{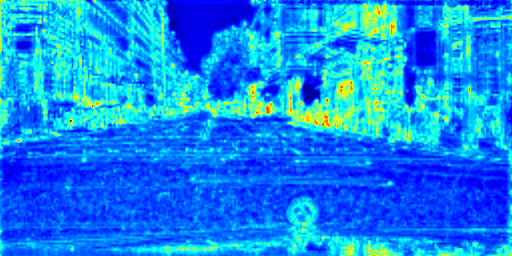}}
  \centerline{(b) $\mathbf{x}_h$}\medskip
\end{minipage}
\begin{minipage}[b]{0.2\linewidth}
  \centering
  \centerline{\includegraphics[width=\linewidth]{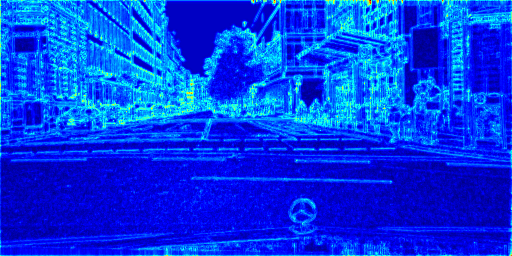}}
  \centerline{(d) $\mathbf{x}_l$}\medskip
\end{minipage}
\begin{minipage}[b]{0.2\linewidth}
  \centering
  \centerline{\includegraphics[width=\linewidth]{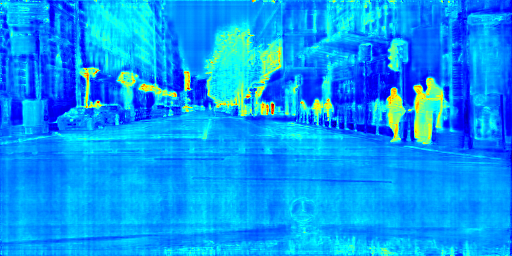}}
  \centerline{(e) $\mathbf{x}_{\mathbf{S}out}$}\medskip
\end{minipage}
\begin{minipage}[b]{0.2\linewidth}
  \centering
  \centerline{\includegraphics[width=\linewidth]{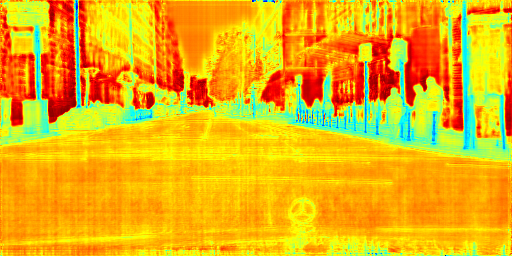}}
  \centerline{(f) $\mathbf{x}_{\mathbf{S}out}+\mathbf{x}_{\mathbf{C}out}$}\medskip
\end{minipage}
}
\caption{\textbf{Visualization of DDSM.} (a) Input, (b) centers of \textcolor{red}{red crosses} indicate 25 spatial positions dynamically sampled for the center of the \textcolor{blue}{blue cross}, (c) the coarse high-level features $\mathbf{x}_h$ with rich semantics, (d) detailed low-level features $\mathbf{x}_l$, (e) output of our spatial-wise module $\mathbf{x}_{\mathbf{S}out}$, with semantics dynamically assigned to detailed features, (f) summation of spatial-wise and channel-wise output, which emphasizes the boundaries. Best view it in color and zoom in. }
\label{fig:visual}

\end{figure*}

In order to verify the effectiveness of our proposed DDSM, we conduct thorough experiments on Cityscapes\cite{Cityscapes} and CamVid\cite{CamVid}. The mean Intersection over Union (mIoU) is adopted as the evaluation metric in all experiments, and F-Score\cite{perazzi2016benchmark} is used to measure the boundary performance.

\noindent
\textbf{Implementation details:} Our method is implemented using the Pytorch framework. For all our experiments, an SGD is used as the optimizer, momentum and weight decay are set to 0.9 and 5e-4, respectively. The learning rate is set as 0.01 and is decayed by multiplying $(1-\frac{\text{epoch}}{\text{max\_epoch}})^{0.9}$. For data augmentation in training, we employ a random horizontal flip, a random resize with scale range [0.75,2], and then a random crop of $1024\times1024$ for Cityscapes ($720\times720$ for Camvid).\\
\noindent
\textbf{Ablation study: }
\label{sec:ablation}
To verify the effectiveness of each component of our method, we conduct multiple sets of experiments on Cityscapes, including whether to use DCN~\cite{deformable}, spatial-wise and channel-wise dynamic affinity modeling or not. We insert two DDSMs into the second and third stages of UPerNet~\cite{upernet}. The number of sampled spatial positions and channels are both set to 9. As shown in Table~\ref{ablation}, based on UPerNet~\cite{upernet}, both spatial and channel-wise dynamic modules will bring more benefits than DCN~\cite{deformable}. A mix of both modules improves the performance by 1.26\%. Simultaneously, to verify the applicability of DDSM in different frameworks, we insert one DDSM in Deeplabv3+~\cite{deeplabv3p}. Table~\ref{ablation}(right) also shows the performance improvement of DDSM 
on Deeplabv3+.

\begin{table}[!t]\setlength{\tabcolsep}{6pt}
	\centering
\caption{\textbf{Ablation Study on Cityscapes} \emph{val} \textbf{set.} \textbf{S}: Spatial-wise dynamic affinity modeling, \textbf{C}: Channel-wise dynamic affinity modeling. All networks use ResNet50 as backbone. And $\Delta$(\%) means the absolute numerical improvement.
}
	\begin{threeparttable}
	\scalebox{0.67}{
\begin{tabular}{@{}lcclcc@{}}
\toprule[0.1em]
\multicolumn{3}{c}{Ablation on UPerNet~\cite{upernet}}                         & \multicolumn{3}{c}{Ablation on Deeplabv3+~\cite{deeplabv3p}} \\ \midrule

Method          & mIoU(\%)  & $\Delta$(\%) & Method           & mIoU(\%)  & $\Delta$(\%) \\ \midrule
UPerNet~\cite{upernet}      & 78.39 & -        & Deeplabv3+~\cite{deeplabv3p}      & 77.76     & -           \\
UPerNet+DCN~\cite{deformable}       &  78.51 & 0.12         & Deeplabv3++DCN~\cite{deformable}  &  78.09     & 0.33        \\
UPerNet+\textbf{S}       &  79.25 & 0.86  & Deeplabv3++\textbf{S}    & 78.34     & 0.58        \\
UPerNet+\textbf{C}     & 78.99 & 0.60        & Deeplabv3++\textbf{C}    & 78.22     & 0.46        \\
UPerNet+\textbf{S}+\textbf{C}& 79.65 & 1.26           & Deeplabv3++\textbf{S}+\textbf{C}  & 78.57     & 0.81       \\
\bottomrule[0.1em]
\end{tabular}
    }

\label{ablation}
\end{threeparttable}
\end{table}

\noindent
\textbf{Analysis of Boundary F-Score and Visualizations:} To show the advantages of our model at the boundaries, we adopt the boundary F-Score~\cite{perazzi2016benchmark} to measure the segmentation accuracy at the boundaries. Table~\ref{tab:f-score} shows the boundary F-Score under different thresholds. Our model is entirely ahead of the baseline, proving its advantages at the boundaries. We also visualize the input and output features of DDSM, as shown in Fig.~\ref{fig:visual}. All feature maps are averaged along channels for display. The visualizations show that DDSM can dynamically propagate high-level semantic information to detailed low-level features from spatial and channel domain.
\begin{table}[!t]\setlength{\tabcolsep}{6pt}
\caption{\textbf{Boundary F-Score on UPerNet~\cite{upernet}.}}
	\centering
	\begin{threeparttable}
	\scalebox{0.70}{
\begin{tabular}{cccccc}
\toprule[0.1em]
Threshhold & 3px  & 5px  & 9px  & 12px & mean \\ \midrule
UPerNet~\cite{upernet}    & 66.4 & 76.3 & 80.1 & 81.5 & 76.1 \\
Ours       & 69.3 & 78.9 & 82.3 & 83.6 & 78.5 \\ \bottomrule[0.1em]
\end{tabular}
}

\label{tab:f-score}
\end{threeparttable}
\end{table}

\noindent
\textbf{Comparison with previous work:}
The mIoU of our results on the Cityscapes test set reaches 81.7\%, which performs favorably against state-of-the-art segmentation methods. Meantime, our method has shown advantages in terms of computational consumption, which is only about 30\% of DAnet's~\cite{DAnet}, as Table~\ref{tab:city-test} shows. We sample 25 pixels and 9 channels here to obtain further performance improvement according to the ablation study. We insert three DDSMs in UPerNet~\cite{upernet} to form $\mathbf{\widetilde x}_4$, $\mathbf{\widetilde x}_3$, $\mathbf{\widetilde x}_2$.  Multi-scale testing is conducted following CCNet~\cite{ccnet}. Table~\ref{tab:city-test} also shows the advantages of our method over the Non-Local based methods.

\begin{table}[!t]\setlength{\tabcolsep}{6pt}
\caption{\textbf{Comparison on Cityscapes} \emph{test} \textbf{set.} Only the methods that merely use the fine dataset are listed. GFlops is measured by 1024 $\times$ 1024 inputs.
}
	\centering
	\begin{threeparttable}
		\scalebox{0.70}{

\begin{tabular}{@{}lccccc@{}}
\toprule[0.1em]
Model     & Reference & Backbone     & mIoU(\%)  & \#Params & \#GFLOPs  \\ \midrule[0.1em]
DFN~\cite{dfn}       & CVPR2018  & ResNet-101   & 79.3  & 90.7M  & 1121.0    \\
PSANet~\cite{psanet}    & ECCV2018  & ResNet-101   & 80.1  & 85.6M  & 1182.6      \\
DenseASPP~\cite{denseaspp} & CVPR2018  & DenseNet-161 & 80.6   & 35.7M & 632.9      \\
ANNet~\cite{anl}      & ICCV2019  & ResNet-101   & 81.3   & 63.0M & 1089.8       \\
CPNet~\cite{cpnet}     & CVPR2020  & ResNet-101   & 81.3   & - & -       \\
CCNet~\cite{ccnet}     & ICCV2019  & ResNet-101   & 81.4 & 66.5M & 1153.9 \\
RGNet~\cite{Yu-ECCV-RepGraph-2020} & ECCV2020 & ResNet-101  & 81.5 & - & - \\
DANet~\cite{DAnet} & CVPR2019 & ResNet-101   & 81.5   & 66.6M & 1298.8 \\
\midrule
Ours      &     -      & ResNet-101   & \textbf{81.7} &  \textbf{51.8M} & \textbf{367.5} \\
\bottomrule[0.1em]
\end{tabular}}

\label{tab:city-test}
\end{threeparttable}
\end{table}

\noindent
\textbf{Experiments on CamVid:}
To further verify the effectiveness of DDSM, we also conduct experiments on the CamVid dataset. Table~\ref{CamVid} shows our results on CamVid. Our model without Cityscapes pre-training outperforms the others. After pre-training, the mIoU performance is improved to 80.6\%.

\begin{table}[!t]\setlength{\tabcolsep}{6pt}
\caption{\textbf{Comparison on CamVid} \emph{test} \textbf{set.} We do not adopt multi-scale testing or other tricks.}
	\centering
	\begin{threeparttable}
		\scalebox{0.7}{
\begin{tabular}{lccc}
\toprule[0.1em]
Method       & Pre-train  & Backbone   & mIoU(\%)      \\ \midrule
PSPNet~\cite{pspnet}       & ImageNet   & ResNet50   & 69.1          \\
DenseDecoder~\cite{densedecoder} & ImageNet   & ResNeXt101 & 70.9          \\
VideoGCRF~\cite{VideoGCRF}    & Cityscapes & ResNet101  & 75.2          \\ \midrule
Ours         & ImageNet   & ResNet101  & \textbf{77.1}          \\ 
Ours         & Cityscapes & ResNet101  & \textbf{80.6} \\
\bottomrule[0.1em]
\end{tabular}
}

\label{CamVid}
\end{threeparttable}
\end{table}

\section{Conclusion}
\label{conclusion}
We propose an end-to-end trainable Dynamic Dual Sampling Module for both spatial-wise dynamic affinity modeling and channel-wise dynamic affinity modeling between two different features. Thus lower layer features are dynamically enhanced by the features of representative pixels and channels from the higher layer simultaneously. Through lots of experiments, our proposed DDSM is verified to be effective on different networks.  Our model achieves advanced performance on the Cityscapes and Camvid datasets while significantly reducing computational consumption.

\small{
\bibliographystyle{IEEEbib}
\bibliography{egbib}
}

\end{document}